\pdfoutput=1

\documentclass[11pt]{article}

\usepackage[final]{acl}

\usepackage{times}
\usepackage{latexsym}

\usepackage[T1]{fontenc}

\usepackage[utf8]{inputenc}

\usepackage{microtype}

\usepackage{inconsolata}

\usepackage{graphicx}

\usepackage{enumitem} 
\usepackage{tabularx}
\usepackage{graphicx}
\usepackage{booktabs}
\usepackage{multirow}
\usepackage{amsmath}
\usepackage{soul}
\usepackage{tcolorbox}
\usepackage{xcolor}
\usepackage{lineno}
\usepackage{enumitem}
\usepackage{subcaption}
\usepackage{tabularx}
\usepackage{xcolor}
\usepackage{colortbl}
\definecolor{lightblue}{rgb}{0.8, 0.93, 1}  
\sethlcolor{lightblue}  
\newcommand{\highlight}[1]{\hl{#1}}  

\title{LLM-Based Robust Product Classification in Commerce and Compliance}

\author{First Author \\
  Affiliation / Address line 1 \\
  Affiliation / Address line 2 \\
  Affiliation / Address line 3 \\
  \texttt{email@domain} \\\And
  Second Author \\
  Affiliation / Address line 1 \\
  Affiliation / Address line 2 \\
  Affiliation / Address line 3 \\
  \texttt{email@domain} \\}

\author{
  \textbf{Sina Gholamian},
  \textbf{Gianfranco Romani},
  \textbf{Bartosz Rudnikowicz},
  \textbf{Stavroula Skylaki}
\\
  Thomson Reuters Artificial Intelligence Labs
\\
  \small{
  {\{sina.gholamian, gianfranco.romani, bartosz.rudnikowicz, laura.skylaki\}@thomsonreuters.com}
 }
}

\begin{document}
\maketitle

\begin{abstract}
Product classification is a crucial task in international trade, as compliance regulations are verified and taxes and duties are applied based on product categories. 
Manual classification of products is time-consuming and error-prone, and the sheer volume of products imported and exported renders the manual process infeasible. 
Consequently, e-commerce platforms and enterprises involved in international trade have turned to automatic product classification using machine learning. 
However, current approaches do not consider the real-world challenges associated with product classification, such as very abbreviated and incomplete product descriptions. 
In addition, recent advancements in generative Large Language Models (LLMs) and their reasoning capabilities are mainly untapped in product classification and e-commerce. 
In this research, we explore the real-life challenges of industrial classification and we propose data perturbations\footnote{We use `data perturbation' and `data attack' interchangeably.} that allow for realistic data simulation. 
Furthermore, we employ LLM-based product classification to improve the robustness of the prediction in presence of incomplete data. 
Our research shows that LLMs with in-context learning outperform the supervised approaches in the clean-data scenario. 
Additionally, we illustrate that LLMs are significantly more robust than the supervised approaches when data attacks are present.   
\end{abstract}

\section{Introduction}
Product classification plays an important role in international trade and e-commerce. 
This is because import and export tariffs are assigned based on the category of products. 
According to the latest report from World Custom Organization~\cite{WCO_website}, in Year 2022-2023 more than 1.3 billion declarations are booked through customs worldwide~\cite{WCOAnnualReport2022}. 
This massive workload, a result of trade globalization, can impose a significant burden on human experts such as customs personnel and the companies involved in import, export, and e-commerce. 

In addition, product classification can often be a complicated task and require subject matter expertise, as there is a wide range of products traded across various industries. 
As such, for human personnel to become competent in understanding the nuances of different products and how to classify them in compliance with WCO guidelines is a non-trivial task and requires several months of training, according to our subject matter expertise. 
Moreover, correct and detailed classification is critical, as incorrect classification can lead to tax liabilities owed to authorities. 
This can result in fines, penalties, and in some cases, legal repercussions and business discontinuation bans in the jurisdictions affected by a tax breach.

Managing the increasing workload of product classification in global trade is difficult. 
This challenge is further compounded by the continuous globalization of e-commerce. 
Additionally, staying accurate and up-to-date as global trade classification guidelines, such as the Harmonized System~\cite{HSsys}, which continuously change, further adds to the challenges of manual product classification.
Therefore, many organizations active in industry have adopted automated methods of product classification using machine learning~\cite{avigdor-etal-2023-consistent,hasson2021category,lee2021classification,chen2021multimodal,nguyen-khatwani-2022-robust}. 
However, the issue with current classification approaches is that they primarily focus on the `clean' version of data, often ignoring the common data perturbations that happen in real-world product classification during inference time. 
In this context, `perturbations' or `attacks' refer to issues in data that limit the classifier's performance, such as incomplete or abbreviated data. 
The ability to robustly predict correct product classifications in scenarios where data might be far from perfect is of paramount importance, especially in cases where incorrect classification can lead to incorrect taxation and trade liabilities in international trade under the harmonized system~\cite{WCOHS}.   
Therefore, in this research, we aim to understand which models perform better in scenarios with potential data attacks. 
This not only facilitates more informed model decision-making, but also considers real-life data challenges.
Consequently, our contributions are as follows:\vspace{-1mm} 
\begin{itemize}
\item We introduce a framework designed to simulate real-life data attacks on clean data. 
This is particularly crucial for product classification with compliance implications, where incorrect classifications can lead to wrong taxation. 
\item Utilizing realistic data attacks, we propose an LLM-based classification approach that outperforms the prior supervised approaches, and is more robust to real-life data attacks. 
\item Lastly, we offer a comprehensive evaluation of human annotators and various models across  different attack scenarios and compare their robustness. 
We draw conclusions from our findings, which we believe are instrumental in guiding design decisions for the practical aspects of product classification.
\end{itemize}

\section{Background}
This section provides a review of the related work and essential background that supports our research.
\subsection{Product Classification}
Product classification based on product description text has been a focal point in several industrial research efforts~\cite{kondadadi2022data,nguyen-khatwani-2022-robust,hasson2021category,avigdor-etal-2023-consistent}. 
In real-world scenarios, product descriptions often lack completeness and in many cases are abbreviated and brief. 
This provides very limited context for accurate product classification using Natural Language Processing (NLP) approaches. 
Kondadadi et al.~\cite{kondadadi2022data} presented a Question Answering (QA) framework for Data Quality Estimation (DQE) with the goal of improving product classification for tax code assignment. 
This approach detects the quality of available data by extracting attribute-value pairs. 
The authors similarly observed that the input product description data is generally vague and noisy.  
Hasson et al.~\cite{hasson2021category} discussed the classification challenges in e-commerce systems. 
Notably, the high diversity of products to classify and highly granular hierarchy of these products result in hundreds or thousands of possible categories, which can present challenges for both manual and automated classification approaches. 
Considering that automated product classification is a more cost-efficient and scalable approach to adopt, the development of robust product classification in presence of data attacks still remains largely unexplored. 
\subsection{Input Perturbation} 
Perturbations in data, specifically in text data, have been investigated in several prior studies~\cite{behjati2019universal,zhang2020adversarial,zou2023universal}. 
Generally, for LLMs, adversarial attacks can involve malicious tokens added to the prompt that causes the model to generate undesired outputs~\cite{zou2023universal}. 
Beyond malicious intents, adversarial attacks can be beneficial and be leveraged as data augmentation to improve the robustness of text classification approaches~\cite{yoo-qi-2021-towards-improving,wang2020infobert,wang-etal-2022-measure} in scenarios where the inference data can be noisy~\cite{morris2020textattack}. 
Our work focuses on product classification based on the text description of products, which in real life can be incomplete and far removed from the clean training data. 
Therefore, in this research, we focus on formulating data perturbations that aim to simulate the real-world data incompleteness often encountered in product descriptions.

\section{Methodology}\label{method}
Although product classification is generally tested on datasets free of inaccuracies, in real-world scenarios the data received from users is often very short and abbreviated.
As such we define an adversarial attack framework to simulate realistic data from clean data. 
For data perturbation method, we follow the approach introduced in~\cite{behjati2019universal}. 
Similar to the method explained in GPT3Mix approach~\cite{yoo-etal-2021-gpt3mix-leveraging}, we use GPT-4 (\textit{version: 0613}) to create perturbations and generate synthetic yet highly realistic datasets to resemble the real-life scenario of the data. 
We write a prompt that includes the instructions to GPT-4 for different variations of data perturbations. 
These instructions are then passed to GPT-4 along the original product description to perform perturbations. 
In response, GPT-4 completion returns the perturbed product description.   
Additional details on prompt templates are provided in Figures~\ref{prompt_data_attack} and~\ref{prompt_classification} in Appendix~\ref{prompts}. 
As outlined in the prompts, we instruct GPT-4 to perform controlled data perturbations so that the initial meaning of the descriptions is still mostly preserved and they remain classifiable by a human annotator.

\subsection{Data Perturbation Framework}
To simulate real-world data scenarios, we introduce realistic data perturbations and attacks. 
Our perturbation model is defined as follows: consider a classifier $f$, which maps an input $x \in X$ to its corresponding class $c\in C$, denoted as $f(x)=c$. 
In this model, $x$ is a sequence of tokens, $x = (x_1,x_2,...,x_n)$. 
Data perturbation can involve either removing or modifying tokens within $x$, leading to a new sequence, $x' = (x_1',x_2',...,x_n')$. 
This perturbation may result in $f(x')=c'$, where $c' \neq c$, indicating a change in classification. 
To mimic the real-life data, we apply two distinct perturbation methods that we will discuss in the following.    

\subsection{Amputation} 
In this approach, we perturb the product description by randomly removing some of its tokens. 
We investigate this scenario because real data often is missing critical attributes, which limits accurate classification of products~\cite{kondadadi2022data}. 
Here, we do not introduce new tokens (i.e., new attributes) nor change the order of the existing tokens; instead we only omit some tokens from the product descriptions. 
Formally speaking, the input $x = (x_1,x_2,...,x_n)$ is transformed into $x_m = (x_{i_1}, x_{i_2}, \ldots, x_{i_k}) \text{ where } 1 \leq i_1 < i_2 < \ldots < i_k \leq n \text{ and } \forall x_{i_{1:k}} \in x$.

\subsection{Abbreviation} 
In this approach, we attack product descriptions by replacing a subset of words with their abbreviated forms. 
This scenario does not fully remove any tokens but converts certain tokens into their abbreviated versions. 
For example, the word \textit{`mobile'} could be replaced by \textit{`mob.'} (refer to Table~\ref{table:example}). 
Formally, the input $x = (x_1,x_2,...,x_n)$ is transformed into $x_a = (x_1', x_2', \ldots, x_n')$ where $S \subseteq \{1, 2, \ldots, n\}$ and $\forall i \in S: x_{i}' = Abbr(x_i)$, and $\forall i \in \{1, 2, \ldots, n\}\setminus S: x_{i}' = x_i$. 

It should be noted that our framework does not encompass a comprehensive list of data perturbation that can happen in real-world scenarios, and only models the common perturbations in our enterprise global trade use case. 
Other data perturbations, such as typos, can also be quite prevalent in real scenarios which can be investigated as per use case. 

\subsection{Example - Data Perturbation} Table~\ref{table:example} provides examples of various attacks based on our data perturbation framework. 
In a combined attack, both abbreviation and amputation approaches are applied.
\begin{table}[h]
\centering
\small
\begin{tabularx}{.48\textwidth}{p{1.4cm}|p{5.5cm}}
\toprule
\textbf{Attack} & \textbf{Description}  \\
\midrule
Clean  & Samsung ALC820 mobile phone case Cover Brown\\
Abbreviated  & samsung alc820 mob. phone case cover brwn\\
Amputated  & samsung alc820 mobile phone case\\
Combined  & samsung alc820 mob. phone case\\
\bottomrule
\end{tabularx}
\caption{Examples of various data attacks applied to clean data.}
\label{table:example}
\vspace*{-4mm}
\end{table}

\subsection{Robustness Metric} 
We define the robustness of classifier $f$  as the delta ($\Delta_r$) of the performance metric ($M$) on the clean data ($D_c$) versus the performance of the classifier on the perturbed data ($D_p$): $\Delta_r(f) = \frac{|M(D_c)-M(D_p)|}{M(D_c)}\vspace{1mm}$. 
The lower the $\Delta_r$, the more robust the model is to the data perturbations.  

\subsection{Research Hypothesis} 
Our hypothesis posits that LLMs with in-context learning not only can outperform supervised models in the product classification task, but also show greater robustness to adversarial attacks such as abbreviation and amputation. 
Furthermore, we assert that informing an LLM about the potential data attacks can improve the classification performance by allowing the LLM to more effectively leverage its reasoning capabilities.


\section{Evaluation}
In the following, we outline the details of our evaluation.
\subsection{Datasets}
We experiment on two publicly available datasets, namely Icecat~\cite{icecat} and WDC-222~\cite{wdc222}, to demonstrate our perturbation framework and evaluate the robustness of different classification models in the presence of data attacks. 
Although we have observed the aforementioned attack scenarios in our proprietary data, we believe our perturbation framework can be readily applied to any arbitrary dataset. 
Therefore, we opt to conduct our evaluation on public datasets to ensure higher visibility and reproducibility. 
The datasets are as follows:

\subsection{Icecat~\cite{icecat}} 
This dataset features products in the ``Computers \& Electronics'' category, organized in a hierarchical structure. 
Each record includes a product description and a corresponding text label. 
For example, as shown in Table~\ref{table:example}, the product described as  \highlight{\textit{``Samsung ALC820 mobile phone case Cover Brown''}} falls under the hierarchy \highlight{\textit{Computers \& Electronics $\rightarrow$ Telecom \& Navigation $\rightarrow$ Mobile Phone Cases}}, with the label being the leaf node of this hierarchy, i.e., \textit{Mobile Phone Cases}. 
The dataset has 370 leaf nodes, with 489,902 entries for training and 153,095 for testing. 
We utilized the entire training set for training supervised models and identifying few-shot examples for LLMs. 
However, to contain LLM inference costs, we conducted stratified random sampling on test set to comprise a smaller set of 5,000 examples, with at least one data point from each class.  

\subsection{WDC-222~\cite{wdc222}} This dataset contains 222 leaf nodes in the same hierarchy as Icecat. 
It includes 2,984 entries solely for testing, thereby serving as the gold standard for this classification task. 
This dataset is generally more difficult than Icecat for classification, and prior approaches~\cite{wdc222} achieve a lower performance on this dataset than Icecat. 
We utilize the entire size of this dataset to test both supervised and large language models.

\subsection{Models}
We conduct our evaluation using both supervised and LLM-based approaches.  

\subsection{Supervised Baseline} To compare the performance of generative models against supervised models, we experiment with the DeBERTaV3-base model~\cite{he2023debertav} as our baseline.
This architecture achieves state-of-the-art performance on several text classification benchmarks.
Specifically, we used the pretrained model available from HuggingFace~\cite{Wolf_Transformers_State-of-the-Art_Natural_2020}, and fine-tuned it on the full training set of the Icecat dataset.
By doing so, we replicate a scenario where the model is trained on clean data and tested on perturbed data, which is a common situation in our real-world use case.  
For the supervised baseline, experiments are repeated several times with different seeds, and thus error ranges are provided. 

\subsection{Training Details}\label{apx_training}
In the following, we review the training details for supervised baseline models.
\subsubsection{Flat Classification}
To train both hierarchical and flat baselines, we used the DeBERTaV3-base model~\cite{he2023debertav}. 
We fine-tuned the pretrained model provided by the authors of the model and available on the Hugging Face~\cite{deberta_v3_huggingface}.
We used the default tokenizer provided by Hugging Face for the DeBERTaV3-base model and the following hyperparameters: batch size of 32, learning rate of 2e-5, and weight decay of 0.01.
The rest of the parameters were equal to default values for the Hugging Face Trainer class. 
We trained the model for a maximum 100 epochs with early stopping enabled and the patience parameter was set to 5 epochs. 
The model was trained on 5 different random seeds, and each converged before reaching the maximum number of epochs.

\subsubsection{Hierarchical Classification}
For the hierarchical classification, we used the same model, tokenizer, and hyperparameters as for the flat classification.
However, we trained two separate models: one with the task to classify the products to the second level of the hierarchy (first level was shared among all products), and the second model for final label prediction. 
The top-level model was trained on the same data as the flat classification model.
The second model was trained on the same data, but the description was augmented with the top-level category label (in textual form) in the following format "\textit{original\_description}, \textit{category\_name}".
During inference, we used predictions from the top-level model and appended them to the description before passing it to the second model for the final classification. 
The results were averaged for the models trained on five different seeds and rounded to three decimal digits. 
We also reported the 95\% interval which was calculated as follows:  $\pm 1.96 \cdot  \frac{std}{\sqrt{5}}$.

\section{LLMs} 
We experiment with both open-source and proprietary LLMs, including Llama 2 Chat with 70B parameters~\cite{touvron2023llama}, GPT3.5, and GPT4 (\textit{model version: 0613})~\cite{openai2023gpt4}. 
Unlike the supervised approach, we were not able to perform multiple runs and report error ranges for LLMs due to the excessive cost of inference. 
However, we set the temperature values to $0$ to minimize potential variations in the LLM outputs across multiple runs. 
\subsection{Models Configurations}
For classification configurations, we consider \textbf{Flat}, \textbf{Hierarchical}, and \textbf{Few-shot} configurations. 
In the flat configuration, the model is tasked with directly predicting the leaf node label of the product, corresponding to 370 and 222 classes for the Icecat and WDC datasets, respectively. 
In the hierarchical configuration, the model initially predicts the second-level hierarchy of the product which is 17 classes for both Icecat and WDC-222 dataset (first-level hierarchy, \textit{Computers \& Electronics}, is shared among all products). 
This is followed by predicting the final leaf label from the predicted second-level hierarchy. 
For the few-shot configuration, we select the top-5 semantically similar examples to the test product from the training set, using the SentenceTransformer model~\cite{reimers-2019-sentence-bert}. 
These examples are then included in the prompt as in-context learning examples for the LLMs~\cite{brown2020language}.
\subsection{Attack Configurations}
We explore four different attack configurations as discussed in our data perturbation framework in Section~\ref{method}. \textbf{Clean:} this configuration presents the original data without any attacks, e.g., the original product descriptions are used for classification. 
This serves as a benchmark for the highest possible classification performance. 
\textbf{Amputated:} in this configuration, the product descriptions are amputated by randomly removing a subset of tokens. 
\textbf{Abbreviated:} this attack involves abbreviating a subset of product description tokens. 
\textbf{Combined:} this configuration involves combining both the amputation and abbreviation attacks, such that the product description is first amputated and then the resulting description is further abbreviated. 
\textbf{Combined-Reason:} this configuration uses the combined attack on the product description, with an additional note in the prompt to enable the LLM to reason about possible data perturbations. 
LLMs have demonstrated emerging capabilities in common-sense reasoning~\cite{wei2022chain}. 
Therefore, in this configuration, we include an extra note in the prompt, \highlight{``Be aware that some parts of the product description might have been abbreviated or amputated.''}, to let the LLM reason on possible perturbation patterns in the product description, which may lead to more accurate classification.
\begin{table}[h]
\centering
\small
\begin{tabular}{c|ccc}
\toprule
\textbf{Similarity} & \textbf{Abbreviated}  &  \textbf{Amputated} & \textbf{Combined} \\
\midrule
Icecat  &$0.918$ &  $0.909$   & $0.848$ \\
WDC-222  & $0.896$  &  $0.907$   & $0.843$ \\
\bottomrule
\end{tabular}
\caption{Similarity scores for the clean dataset versus the attacked datasets.}
\label{table:similarity}
\vspace*{-4mm}
\end{table}

\subsection{Data Analysis}\label{data_analysis}
\begin{figure}[ht]
\centering
\begin{subfigure}{0.235\textwidth} 
        \centering
        \includegraphics[width=\textwidth]{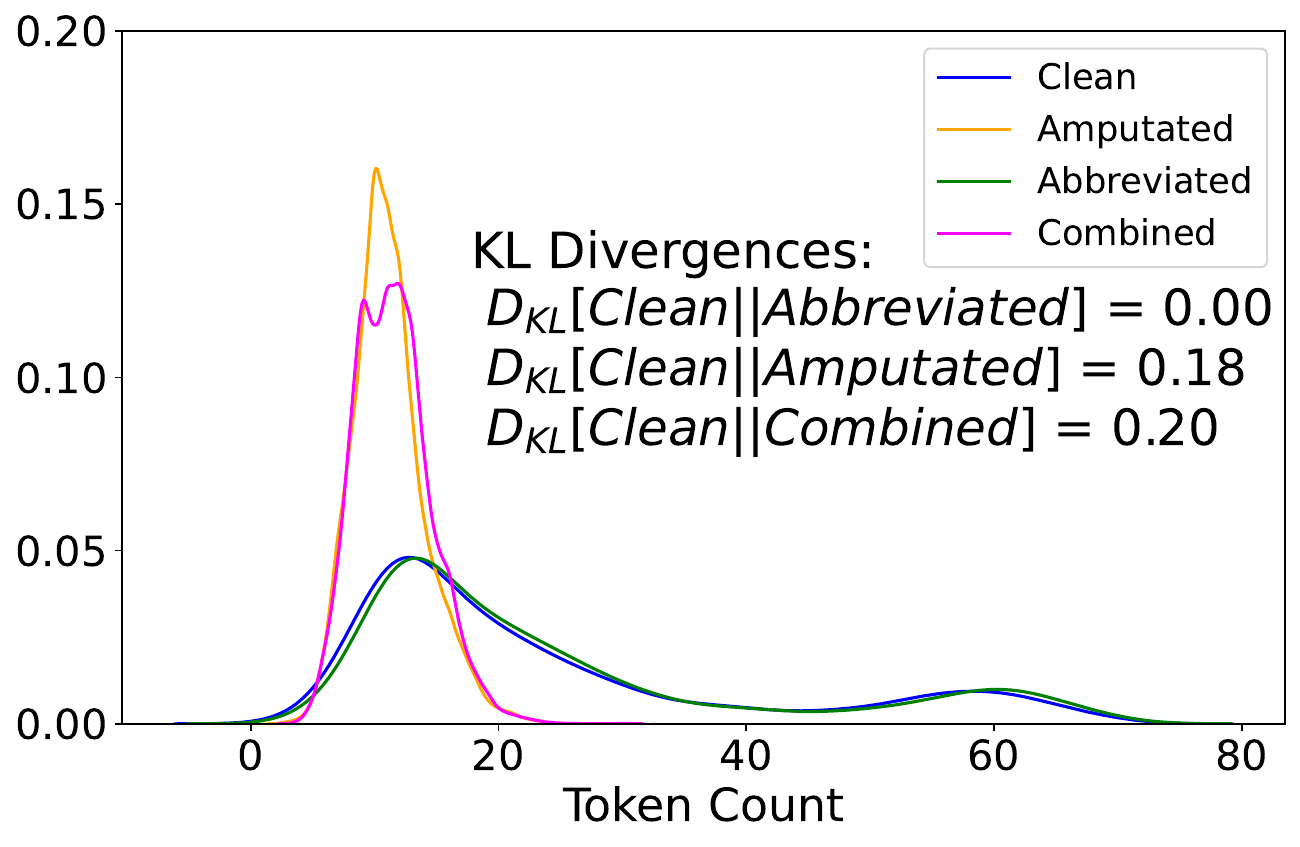}
        \caption{Icecat}
        \label{fig:part_a}
    \end{subfigure}
    \begin{subfigure}{0.235\textwidth}
        \centering
        \includegraphics[width=\textwidth]{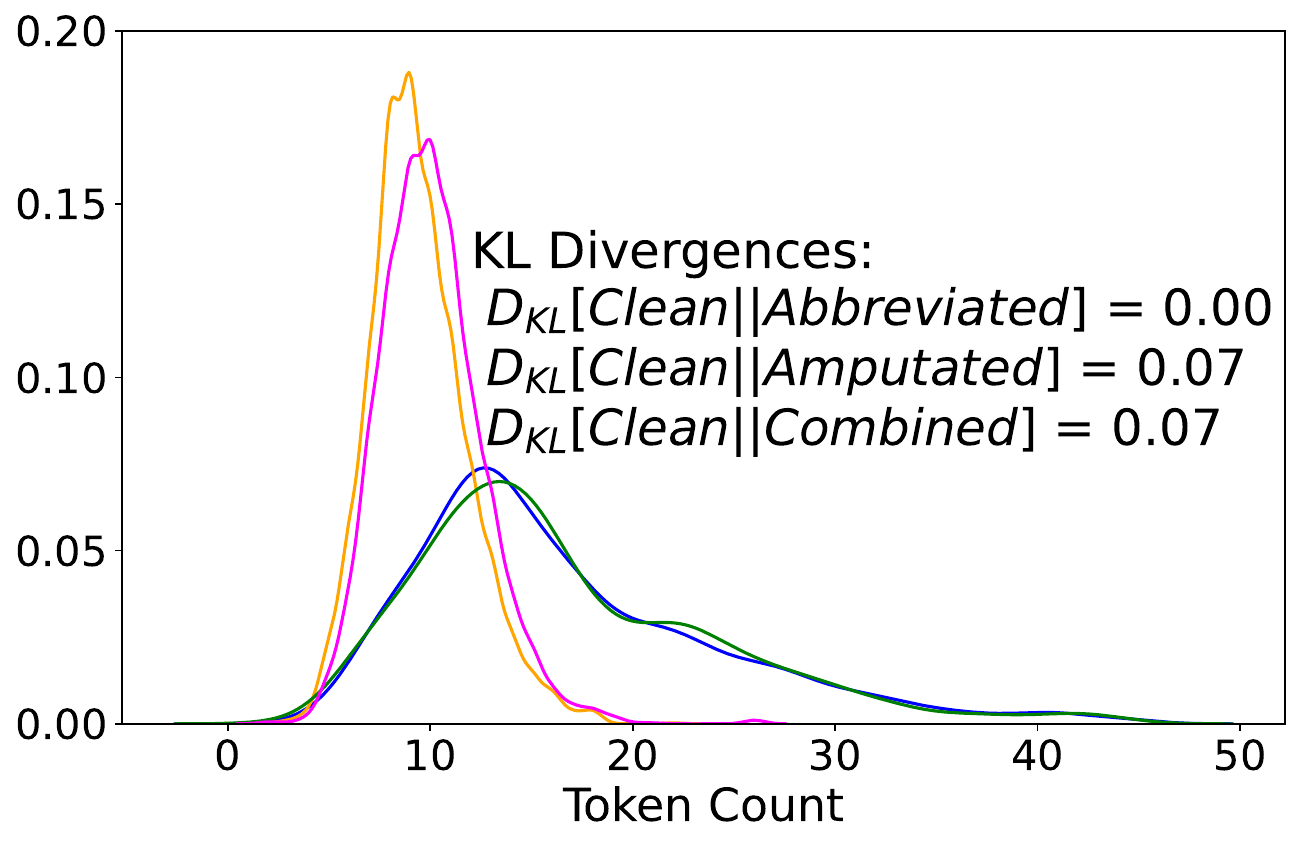}
        \caption{WDC-222}
        \label{fig:part_b}
    \end{subfigure}
    \vspace{-2mm}
    \caption{Distribution of the clean data versus the distribution of the data with different type of attacks.}
    \label{fig:data_dist}
\vspace*{-3mm}
\end{figure}

In this section, we present a statistical analysis of the data attributes for the clean data as compared to the post-attack scenarios. 
Table~\ref{table:similarity} shows the average semantic similarity scores for both the clean dataset and its perturbed ones. 
We used \textit{`multi-qa-mpnet-base-dot-v1'} model from SentenceTransformers~\cite{reimers-2019-sentence-bert} to calculate these similarity scores. 
The results show that as more attacks are introduced on the dataset, the similarity scores decrease. 
However, even for the `Combined' configuration, the dataset is still over $84\%$ similar to the original dataset. 
In addition to the similarity scores, we have plotted the distribution of token sizes for product descriptions in Figure~\ref{fig:data_dist} for both the Icecat (\ref{fig:part_a}) and WDC-222 (\ref{fig:part_b}) datasets. 
Kullback-Leibler (KL) divergence values~\cite{kullback1951information} are also provided for different data configurations. 
Across all configurations, the KL values are less than or equal to $0.2$, and a value of $\le 0.2$ typically signifies a small divergence between the distributions. 
This analysis is crucial as we later evaluate how these small divergences in distributions translates to a greater scale of model performance unrobustness.

\subsection{Human Annotation Analysis}
The importance of the quality of perturbed data prompted us to engage human annotators to assess the quality and ensure its similarity to the intended real data. 
During the design of the data perturbation framework, we leveraged human expert knowledge to ensure our perturbations aligned with the in-field data.  
In addition, through human manual evaluation, we confirmed that the perturbed data appears realistic and plausible in real-life scenarios.  

To further solidify the data quality analysis, we picked 100 random sample data points from each dataset (200 samples in total) that were perturbed and asked our human annotators to expand the abbreviated words to ensure the majority of perturbations are recoverable from a human perspective and they did not semantically alter the meaning of product descriptions. 
Through this, annotators were able to identify and create the clean full form of the abbreviated tokens in the product descriptions 80\% and 85\% of times for the Icecat and WDC-222 datasets, respectively.   

To evaluate that the perturbation process did not semantically alter the descriptions in a significant way, we asked the annotators to label the descriptions with clean descriptions and also combined attack for both datasets (`\textbf{Clean}' and `\textbf{Combined}' in Table~\ref{table:sme}). 
Furthermore, to check if historical classifications of clean descriptions semantically similar to perturbed data would aid annotators, for each combined attack description in the set of 100 randomly selected product descriptions, we provided five most semantically similar examples, using SentenceTransformer~\cite{reimers-2019-sentence-bert} (`\textbf{Combined+FS}' in Table~\ref{table:sme}). 
We then asked the annotator to map the description that is attacked with combined perturbation to its closest clean description. 
Then we calculate the accuracy of the annotator mapped labels versus the true label of the perturbed data points. 
The design for this experiment is similar to adding few-shot similar examples to the LLM prompt to allow the model to find semantic similarities between the original clean data and the perturbed data. 

\begin{table}[h]
\centering
\small
\begin{tabular}{c|ccc}
\toprule
\textbf{Accuracy ($\%$)} & \textbf{Clean}  &  \textbf{Combined} & \textbf{Combined+FS} \\
\midrule
Icecat  &$76$ &  $72$   & $97$ \\
WDC-222  & $72$  &  $67$   & $95$ \\
\bottomrule
\end{tabular}
\caption{Human annotator analysis of perturbed data.}
\label{table:sme}
\vspace*{-2mm}
\end{table}

Table~\ref{table:sme} summarizes the human annotators' classification accuracy results. 
We observe that for both datasets, the combined attack has an impact on the accuracy of classification compared to clean descriptions. 
However, given that we observe high accuracy for both datasets when a few shot semantically similar examples are provided to the annotator, this confirms that the amputation perturbation makes the classification more difficult, but the semantics of the products stay intact. 
This establishes that our perturbation framework works as expected and a classification model that is robust to input perturbations should be able to maintain robust classification performance in the presence of data attacks proposed through our work. 
In the following, we continue with evaluation of machine-learning-based approaches.    

\begin{table*}[t!]
\newcommand{\tuple}[2]{#1 {\small{(#2)}}}
\newcommand{\besttuple}[2]{\textbf{#1} {\small{(#2)}}}

\definecolor{cadmiumgreen}{rgb}{0.0, 0.42, 0.24}
\definecolor{ao(english)}{rgb}{0.0, 0.5, 0.0}
\resizebox{\textwidth}{!}{
\begin{tabular}{ccc|cccccc|ccccccccc}
\toprule
\multirow{2}{*}{\textbf{Model}}                                               &                        &  & \multicolumn{5}{c}{\textbf{Icecat (\%)~\cite{icecat}}} &  & \multicolumn{5}{c}{\textbf{WDC-222 (\%)~\cite{wdc222}}}                                                                      \\ 
\cmidrule(lr){4-9} \cmidrule(l){10-15} 
                                                                              & \textbf{Approach}              &   \textbf{Attack}           & ma-P            & ma-R            & ma-F1             & we-P             & we-R   &we-F1 
& ma-P            & ma-R            & ma-F1             & we-P             & we-R & we-F1\\ \midrule

\multirow{8}{*}{\textbf{\begin{tabular}[c]{@{}c@{}}DeBERTaV3 base\\(Supervised)\end{tabular}}}
& \multirow{4}{*}{\textbf{Flat}}
& Clean  & $88.5 \pm 0.6 $ & $89.2 \pm 0.4$	& $88.3 \pm 0.5$	& $97.9 \pm 0.1$	& $98.1 \pm 0.1$	& $97.8 \pm 0.1$ & $38.9 \pm 2.3$ &	$38.6 \pm 1.7$	& $35.1 \pm 1.8$ & $81.5 \pm 0.8$	& $70.7 \pm 1.4$ & $72.9 \pm 1.5$ \\
& & Abbreviated & $48.1 \pm 1.3$ &	$48.0 \pm 2.1$ & $44.4 \pm 1.7$ & $81.4 \pm 1.6$& $76.0 \pm 3.1$	& $75.8 \pm 2.3$ 
& $25.5 \pm 1.3$	& $21.8 \pm 1.2$	& $19.4 \pm 0.8 $	& $69.2 \pm 2.3$	& $38.4 \pm 3.8$	& $43.6 \pm 4.0$\\
& & Amputated & $67.6 \pm 0.9$	& $72.0 \pm 0.5$	& $67.0 \pm 0.7$	& $87.4 \pm 0.2$	& $85.6 \pm 0.6$	& $85.1 \pm 0.6$ &  
$35.0 \pm 1.9$	& $34.7 \pm 1.4$	& $31.2 \pm 1.6$	& $78.4 \pm 1.4$	& $63.0 \pm 4.3$	& $66.6 \pm 3.8$ \\
& & Combined & $46.0 \pm 0.6$	& $45.9 \pm 1.5$	& $41.7 \pm 0.9$	& $76.2 \pm 0.6$	& $66.7 \pm 2.9$	& $67.6 \pm 2.0$ & 
 $26.0 \pm 1.0$	& $22.0 \pm 1.4$	& $19.7 \pm 0.7$	& $70.5 \pm 0.6$	& $39.9 \pm 3.5$	& $46.0 \pm 3.3$ \\
 
 \cmidrule{2-15}

 & \multirow{4}{*}{\textbf{Hierarchical}}
& Clean  & $83.5 \pm 10.4 $ & $84.8 \pm 9.1$	& $83.4	 \pm 10.0$	& $97.1	 \pm 1.5$	& $97.5 \pm 1.0$	& $97.2 \pm 1.3$ &
 $38.6 \pm 1.4$ &	$37.9 \pm 1.5$	& $34.4 \pm 1.3$ & $81.8 \pm 0.9$	& $68.7 \pm 0.9$ & $71.8 \pm 0.5$ \\
& & Abbreviated & $46.0 \pm 4.5$ &	$46.4 \pm 3.1 $ & $42.4 \pm 3.7$ & $81.2 \pm 1.1$& $73.2 \pm 3.8$	& $74.0	 \pm 2.5$ 
& $26.1 \pm 0.8$	& $22.7 \pm 1.1$	& $20.1	 \pm 0.7 $	& $71.3 \pm 0.9$	& $39.9 \pm 4.6	$	& $45.6 \pm 5.7$\\
& & Amputated & $62.7 \pm 6.8$	& $66.8 \pm 6.3$	& $61.7 \pm 6.3$	& $86.7 \pm 1.2$	& $83.6 \pm 0.6$	& $83.6 \pm 0.7$ &  
$ 36.8	 \pm 1.3$	& $35.6 \pm 1.3	$	& $32.1 \pm 1.1$	& $79.2	 \pm 1.2$	& $60.9	 \pm 2.3$	& $65.2	 \pm 1.5$ \\
& & Combined & $43.2 \pm 4.7$	& $43.3 \pm 3.5$	& $39.0 \pm 3.8$	& $76.0 \pm 1.4$	& $62.4 \pm 3.7$	& $64.8 \pm 2.5$ & 
 $27.0	 \pm 0.7$	& $23.2 \pm 0.8$	& $20.6 \pm 0.5$	& $71.3 \pm 1.3	$	& $41.6 \pm 1.9$	& $47.7 \pm 1.8$ \\

\cmidrule{2-15}
& \cellcolor{gray!15}$\Delta_r$ (\%) & \cellcolor{gray!15}$-$ & \cellcolor{gray!15}$48.0$ & \cellcolor{gray!15}$48.5$ & \cellcolor{gray!15}$52.8$ & \cellcolor{gray!15}$22.2$ & \cellcolor{gray!15}$32.0$ & \cellcolor{gray!15}$30.9$ 
& \cellcolor{gray!15}$33.2$ &  \cellcolor{gray!15}$43.0$ & \cellcolor{gray!15}$43.9$  &  \cellcolor{gray!15}$13.5$ &   \cellcolor{gray!15}$43.6$   & \cellcolor{gray!15}$36.9$\\

\midrule


\multirow{13}{*}{\textbf{\begin{tabular}[c]{@{}c@{}}Llama-2\\(70b-chat)\end{tabular}}} 
& \multirow{4}{*}{\textbf{Flat}}
& Clean  & $19.6$ & $29.2$& $19.9$ & $50.2$ & $37.4$ & $36.9$ 
& $23.8$ & $28.7$ & $21.9$ & $75.9$ & $51.4$& $51.4$\\
& & Abbreviated &   $11.7$ &   $16.8$ & $11.7$ & $78.0$ & $39.4$ & $41.0$
 & $22.5$ & $27.4$ & $20.5$ & $72.5$ & $44.5$ & $42.8$\\
& & Amputated & $16.1$ & $21.6$ & $15.4$ & $81.8$ & $38.3$ & $41.6$ 
&  $25.6$ & $28.2$  & $22.8$ & $76.4$ & $53.4$ & $52.9$\\   
& & Combined &  $13.4$  & $19.5$ & $13.1$& $76.7$  & $40.9$   & $42.0$  
&  $22.6$ &   $27.9$ &    $20.3$ &  $73.3$ & $48.7$ & $47.8$\\
& & Combined-Reason &  $19.9$ & $27.1$ & $19.4$ & $72.2$ & $54.3$ & $54.7$
& $31.0$ & $34.2$ & $27.8$ & $68.7$ & $56.2$ & $52.2$ \\

\cmidrule{2-15}

& \multirow{2}{*}{\textbf{Hierarchical}}
& Clean  & $35.2$ & $34.7$ & $29.8$ & $65.2$ & $40.4$ & $39.4$  
&$33.2$ & $35.7$ & $29.1$& $68.6$ & $41.9$ & $38.1$\\
& & Combined & $32.1$ & $33.6$ & $28.2$ & $58.5$ & $38.5$ & $35.4$ 
& $29.6$ & $32.6$ & $25.4$ & $70.0$ & $37.6$ & $36.8$ \\

\cmidrule{2-15}

& \multirow{4}{*}{\textbf{Few-shot}}
& Clean  & $89.6$ & $89.2$ & $88.3$ & $97.1$ & $96.1$ & $95.9$ 
& $73.1$ & $71.5$ &  $69.4$ & $89.8$ & $86.6$& $85.6$   \\
& & Abbreviated & $76.5$ & $79.0$ & $75.7$ & $85.8$ & $84.5$ & $80.6$ 
& $61.3$ &$67.0$&$59.2$ & $83.8$ &$65.6$&$61.6$\\
& & Amputated & $86.9$ & $85.5$ & $84.8$ & $94.9$ & $93.5$ & $93.1$ 
&   $68.0$ & $68.1$ & $64.3$ & $84.3$& $78.0$ & $74.5$  \\
& & Combined &  $79.3$ & $79.6$ & $77.6$ & $92.7$ & $90.5$ & $89.6$
 & $61.8$ & $65.2$ & $59.2$ & $82.8$ & $68.6$ & $64.5$ \\
& & Combined-Reason & $78.3$ & $78.4$ & $76.3$ & $94.2$ & $92.6$ & $92.6$ 
& $63.7$ & $62.9$ & $59.1$ & $83.0$ & $74.7$ & $72.1$\\

\cmidrule{2-15}

& \cellcolor{gray!15} $\Delta_r$ (\%) &\cellcolor{gray!15} $-$ & \cellcolor{gray!15}$12.6$ & \cellcolor{gray!15}$12.1$ & \cellcolor{gray!15}$13.6$ & \cellcolor{gray!15}$3.0$ & \cellcolor{gray!15}$3.6$ & \cellcolor{gray!15}$3.4$ & \cellcolor{gray!15} $12.9$ & \cellcolor{gray!15}$12.0$  &  \cellcolor{gray!15}$14.8$ &   \cellcolor{gray!15}$7.6$   & \cellcolor{gray!15}$13.7$ & \cellcolor{gray!15}$15.8$\\  

\midrule

\multirow{11}{*}{\textbf{\begin{tabular}[c]{@{}c@{}}GPT3.5\\(ver.: 0613)\end{tabular}}} 
& \multirow{4}{*}{\textbf{Flat}}
& Clean  & $63.9$  &  $63.9$&  $61.0$& $90.4$ &$83.9$ & $84.4$ & $57.1$ & $55.0$   & $53.3$ & $92.2$ & $86.5$& $87.9$\\
& & Abbreviated &   $57.8$ &   $58.6$ &    $54.9$ & $90.0$ &  $82.8$ & $83.5$ &  $54.9$ &   $53.2$   & $51.1$ & $91.2$ &  $85.0$ &  $86.4$  \\
& & Amputated & $64.1$ & $63.8$ & $61.1$ & $89.9$ & $84.3$ & $84.7$ &  $55.5$ & $55.0$  &  $52.5$ &   $90.5$   & $85.1$ & $86.1$\\   
& & Combined &  $57.1$  &  $58.2$    &$54.4$& $88.6$    & $81.6$   & $82.4$  &  $54.9$ &   $53.5$ &    $50.8$ &  $88.2$ &   $82.8$ & $83.2$\\
\cmidrule{2-15}

& \multirow{2}{*}{\textbf{Hierarchical}}
& Clean  &  $63.8$    &$59.0$  &  $57.3$ &  $88.1$ &    $66.0$    &$66.1$  &$58.0$&$53.6$&$51.4$&$81.7$&$65.3$&$66.2$\\
& & Combined & $58.1$ &  $54.2$& $52.1$& $85.8$& $62.8$&  $63.3$ &  $56.5$& $52.5$& $50.0$& $85.7$& $78.5$& $79.0$ \\
\cmidrule{2-15}

& \multirow{4}{*}{\textbf{Few-shot}}
& Clean  & $87.6$ &    $88.3$ &    $87.0$ &  $97.7$ &    $96.7$  &  $97.0$  & $77.0$ &   $76.9$ &    $75.1$ &  $94.1$ &   $92.3$&    $92.5$   \\
& & Abbreviated & $82.5$ & $83.3$ & $81.5$ & $96.7$ & $95.2$ & $95.6$ & $72.0$ &$70.8$&$69.5$&$92.4$&$90.1$&$90.3$\\
& & Amputated & $85.5$ & $85.9$ & $84.6$ & $96.3$ & $95.2$ & $95.4$ 
&  $76.5$  &  $75.7$ &    $74.1$ & $92.7$&  $90.7$ &    $90.8$  \\
& & Combined &  $81.1$   & $82.7$   & $80.1$  & $95.1$   & $93.6$  &  $93.9$ & $72.8$  &  $72.1$ &    $70.0$ & $90.6$  &  $88.1$ &  $87.9$ \\
& & Combined-Reason & $81.3$ & $82.4$ & $80.2$ & $95.4$ & $93.9$ & $94.2$
 &  $72.9$ & $72.4$ & $70.4$ & $89.8$ & $87.3$ & $87.0$\\ 

\cmidrule{2-15}
& \cellcolor{gray!15} $\Delta_r$ (\%) & \cellcolor{gray!15} $-$ & \cellcolor{gray!15}$7.2$ & \cellcolor{gray!15}$6.7$ & \cellcolor{gray!15}$7.8$ & \cellcolor{gray!15}$2.4$ & \cellcolor{gray!15}$2.9$ & \cellcolor{gray!15}$2.9$ & \cellcolor{gray!15}$5.3$ & \cellcolor{gray!15} $5.9$ & \cellcolor{gray!15}$6.3$  &  \cellcolor{gray!15}$4.6$ &  \cellcolor{gray!15} $5.4$   & \cellcolor{gray!15}$5.9$ \\

\midrule

\multirow{9}{*}{\textbf{\begin{tabular}[c]{@{}c@{}}GPT4\\(ver.: 0613)\end{tabular}}}
& \multirow{4}{*}{\textbf{Flat}}

& Clean  & $79.5$& $79.5$  &  $77.5$ & $93.6$& $90.6$& $90.8$&$69.2$& $67.7$  &  $66.0$ & $94.6$ & $89.0$ & $89.9$   \\
& & Combined & $72.9$  &  $73.9$& $71.0$&$92.9$&  $89.9$& $90.2$ & $66.0$ &  $65.6$ &   $63.1$  & $93.3$  &  $88.4$   & $89.1$  \\
& & Combined-Reasoned & $73.6$ & $74.5$ & $71.7$ & $92.8$ & $90.2$  & $90.5$  & $66.8$ &  $66.1$ &   $63.6$  & $93.1$  &  $88.8$   & $89.3$  \\
\cmidrule{2-15}

& \multirow{2}{*}{\textbf{Hierarchical}}
& No-attach  & $66.3$ &$62.1$&$60.8$ & $88.8$&$69.7$&$69.8$ &  $59.4$& $57.4$ &$54.7$&$85.3$&$80.3$&$80.1$\\
& & Combined & $64.1$  &  $59.0$ &   $57.8$ & $81.1$   & $71.9$ &   $69.9$  &  $68.1$  &  $62.2$ & $61.6$&  $87.8$ &  $68.5$   & $68.4$ \\
\cmidrule{2-15}

& \multirow{4}{*}{\textbf{Few-shot}}
& Clean  & ${93.5}$ & ${93.0}$ & ${92.8}$ & $99.0$& $98.5$& $98.6$ & $80.0$&  $77.1$ & $76.9$ & $95.9$  & $94.0$  &  $94.4$ \\
& & Combined & $85.7$ &  $86.2$ &   $84.9$ & $\mathbf{96.9}$  &  $\mathbf{96.0}$ &   $\mathbf{96.2}$ &   $78.0$ &   $76.2$  &  $75.3$ &  $93.8$ &   $91.9$ &  $92.1$  \\
& & Combined-Reason & $\mathbf{86.2}$  & $\mathbf{86.3}$  & $\mathbf{85.2}$  & $\mathbf{96.9}$  & $\mathbf{96.0}$   & $\mathbf{96.2}$  &  $\mathbf{78.7}$ &   $\mathbf{76.9}$ & $\mathbf{75.9}$ & $\mathbf{93.9}$ & $\mathbf{92.1}$ &  $\mathbf{92.2}$  \\
\cmidrule{2-15}
& \cellcolor{gray!15}$\Delta_r$ (\%) & \cellcolor{gray!15}$-$ & \cellcolor{gray!15}$7.8$ & \cellcolor{gray!15}$7.2$ & \cellcolor{gray!15}$8.2$& \cellcolor{gray!15}$2.1$ & \cellcolor{gray!15}$2.5$ & \cellcolor{gray!15}$2.4$ 
 &  \cellcolor{gray!15}$1.6$ & \cellcolor{gray!15}$0.3$  &  \cellcolor{gray!15}$1.3$ & \cellcolor{gray!15}$2.1$ & \cellcolor{gray!15}$2.0$ & \cellcolor{gray!15}$2.3$\\  

\bottomrule
\end{tabular}
}\vspace{1mm}
\caption{The table summarizes the results for Icecat and WDC-222 datasets and different models. We experimented with supervised and large language models for different configurations and attack scenarios. The prefixes ma- and we- denote macro and weighted metrics, respectively. P, R, and F1 denote Precision, Recall, and F1-Score respectively. For each model, the $\Delta_r$ values are calculated for best performing configuration with attacks, i.e., Flat/Combined for supervised and Few-shot/Combined-Reason for LLMs. For each metric, the best-performing configuration with combined data attacks is shown in bold. Note: we-R is comparable to accuracy~\cite{sklearnrecallscore}.}
\label{tab:results}
\vspace{-2mm}
\end{table*}

\subsection{Metrics}
We assess the classification performance using both macro (\textit{ma}) and weighted (\textit{we}) Precision, Recall, and F1-Score values to compare different approaches. 
Additionally, for each model, we also calculate its most robust (i.e., the smallest) $\Delta_r$ score.


\subsection{Robustness Analysis}
Table~\ref{tab:results} shows the performance and robustness of various  configurations that were experimented with. 
It should be noted that we chose to exclude certain configurations from execution in order to manage the models inference API cost and also because we were able to extract patterns from the configurations that were executed. 
We summarize the key observations from the results as follows. 
GPT-4 model with few-shot prompting delivers the best classification results on both datasets among all models and shows the highest level of robustness to the introduced data attacks. 
As expected, the `Clean' data approach yields the best results, with performance marginally decreasing as data attacks are introduced for `Amputated' and `Abbreviated' data configurations. 
Supervised model achieved the second highest performance after GPT-4 for the `Clean' scenario. However, the performance values for this model significantly drop as the attacks are introduced. 
In general, LLMs show more robustness to the introduced attacks in the product description as they are able to better reason on the details of the product description. In addition, few-shot examples allow LLMs to further learn from the context and improve their performance, compared to our experimented supervised classification models which cannot leverage this capability.

Hierarchical classification generally performed equally or worse than flat classification and inferior to few-shot prompting. 
We rationalize that since the errors from the first level of classification propagate to the second level, this compounding effect results in lower performance in hierarchical classification compared to flat configuration. 
In some cases, we observed that hierarchical classification improves macro scores, which indicates that this method achieves a more balanced prediction across different classes. 
For example, Llama-2 achieves better results with hierarchical classification than with the flat classification method.
This is because the hierarchical approach allows the model to focus on a smaller set of classes at each hierarchy.

Comparing the results for two different datasets, Icecat and WDC-222, we observe that LLM-based approaches show a significant improvement for the WDC-222 dataset but a less noticeable improvement for Icecat. 
The reason is that the classification of the Icecat dataset is simpler than that of WDC-222, as the latter comes from heterogeneous data sources~\cite{wdc222}. 
As such, the baseline supervised values for the Icecat dataset are also higher than those for the WDC-222 dataset. 
This also provides grounds for our observation that SOTA LLMs can generalize better than supervised approaches on heterogeneous datasets, based on the noteacible improvement observed in the WDC-222 dataset.

The Few-shot scenario further improves the performance of the LLMs, and GPT-4 achieves a new state-of-the-art result on classification task on Icecat and WDC-222 datasets~\cite{wdc222,brinkmann2021improving}. 
Additionally, the \textbf{`Combined-Reason'} scenario improves classification performance in cases where a combined attack is present.
This added reasoning in the prompt allows to recover some of the performance loss observed between clean data and combined-attack configurations by further leveraging the reasoning capabilities of LLMs. 
Our findings suggest that while LLMs are more robust in classification compared to supervised approaches, i.e., have lower $\Delta_r$s, this \textbf{robustness can be further improved with informing the model of potential data issues, such as missing characteristics and abbreviations.} 
This observation also underlines the need for more practical designs of ML approaches while considering real-world challenges.

\section{Discussion}

\subsection{Data Leakage} One concern that exists is that the LLMs' training dataset, like GPT-4 as an example, might have already included our experimented datasets. 
Although this cannot be entirely ruled out, our approach is still valid for two key reasons. 
Firstly, GPT-4 initially shows lower performance, but significantly improves in our few-shot scenario, outperforming the supervised models. 
This indicates that the effectiveness of GPT-4 extends beyond merely memorization. 
Secondly, the robustness of LLMs, particularly in our data perturbation framework with Combined-Reason, is evident. 
The perturbed dataset, as it is novel and not included in prior training, shows GPT-4’s ability to understand product semantics and effectively recover from data perturbations.

\subsection{Impact and Deployment}
Our research has partially enabled AI-based product categorization in our global trade service which is crucial and sensitive for compliance and regulatory aspects for large corporations active in cross-border trade. Our research is impactful as it has enabled more efficient and accurate classification, and thus reduces the regulatory and compliance risk.
The discovery phase of the project has been completed with testing on millions of data records and the second phase of the project which expands to multiple users and more data is ongoing.

\section{Conclusion}
In this research, we presented a data perturbation framework to simulate the real-world data deficiencies for ML-based product classification. 
We then proceeded with a comprehensive evaluation of different supervised and LLM-based classification approaches in presence and absence of data attacks. 
Our findings show that LLM-based approaches are generally more robust against adversarial attacks and more suitable for applications that require high robustness in predictions and misclassification can cause compliance repercussions. 
As future work, we will further investigate the security robustness of LLMs in data-critical applications and explore leveraging LLMs for providing classification rationales in addition to label predictions. 





\section{Limitations} 
Our analysis has limitations, particularly as we observed that the results from Llama-2, are not completely stable, and small variations within the prompt can lead to noticeable changes in classification performance. 
We believe these limitations are largely addressed in SOTA models, like GPT-4. Additionally, our data perturbation framework models a limited set of data attacks that are relevant to our industrial use case, however, other use cases might face different data challenges, which should be dealt with per use case. 

\section{Ethical and Practical Considerations} 
This study has been carried out by following the privacy requirements of our organization. 
The research has been reviewed by research directors and legal counsel to ensure adherence to privacy of our users data and information. 
Furthermore, the authors of this work have been committed to adhering to the highest standards of ethical responsibility throughout the research.
In product environments where automated product classification models are deployed, the predictions are presented to the end user as suggestions, and it is then the end user's sole responsibility to accept, reject, or manually adjust these predictions as necessary. 
This work presents a general perspective on the product classification task and does not incorporate additional sources of information that could be leveraged for specific use cases, such as the Harmonized System classification, which utilizes tariff schedules, rulings, and keywords.
\bibliography{custom}

\appendix

\section{Prompts}\label{prompts}
Figure~\ref{prompt_data_attack} shows the prompt for simulating data attacks with the help of GPT-4, as explained in the data perturbation framework, while Figure~\ref{prompt_classification} displays the prompt for the classification of products. 
The first prompt aims to is to accurately automate the data perturbation framework, and the second prompt allows to classify the products, using an LLM. 
As the data is manipulated by an LLM, we investigate the correctness of the approach in comparison to the intended outcomes through human analysis in Section~\ref{data_analysis}. 

\begin{figure*}[h]
\begin{tcolorbox}[left=1pt,right=1pt,colback=blue!5!white,colframe=blue!50!white]

\textcolor{gray}{\small (Abbreviation)} You got a new job as a product classifier for products belonging to the Icecat catalog.\\ 
You are asked to modify a description of a product that belongs to the "\{industry\_input\}" category (according to the hierarchy in Icecat) and modify words with their abbreviations (as could happen in shipment details).\\
It is vital to not modify the description in a way that could change the classification of the product.\\ 
Please do not abbreviate more than 20\% of the words or I would not understand the description.\\ 
The order of the words must not change.\\
Original description: \{description\_input\}\\
New description:
\\\\

\textcolor{gray}{\small (Amputation)} You got a new job as a product classifier for products belonging to the Icecat catalog.\\ 
You are asked to truncate a description of a product that belongs to the "\{industry\_input\}" category (according to the hierarchy in Icecat) and to make it much shorter, like it would appear in a shipment detail description.\\ 
Omit all the information that is not strictly necessary to identify the product, i.e. technical characteristics.\\
The order of the words must not change.\\
Work following the order below:\\
1. if the description is shorter than 5 words, do not change it\\
2. if the description is longer than 5 words, select the 5 most important words\\
3. put the selected words in the relative order in which they appeared in the original description\\
Original description: \{description\_input\}\\
New description:
\end{tcolorbox}
\vspace{-2mm}
\caption{This figure shows the prompts used for GPT-4 to perform abbreviation and amputation data attacks.}
\label{prompt_data_attack}
\end{figure*}

\begin{figure*}[h]
\centering
\begin{tcolorbox}[width=0.95\textwidth, left=1pt,right=1pt,colback=blue!5!white,colframe=blue!50!white]
\setlength\linenumbersep{.4cm} 
\setcounter{linenumber}{1}
\renewcommand{\linenumberfont}{\normalfont\small\sffamily\color{black}}
\renewcommand{\thelinenumber}{\ifnum\value{linenumber}<10 0\fi\arabic{linenumber}} 
\begin{internallinenumbers}
\textbf{Classify the following product to one class form the list below.\\}

\textbf{List of classes:\\} 
\textit{Warranty \& Support Extensions}\\
\textit{Notebooks}\\
\textit{PCs/Workstations}\\
\textit{...}\\

\textcolor{gray}{\small(Few-shot)} \textbf{Some examples with their classes are provided:}\\
\hspace{5mm}\{\textit{5-shot similar examples}\}\\

\textbf{Product:} \{\textit{test product}\}

\textcolor{gray}{\small (Combined-Reason)} Be aware that some parts of the product description might have been abbreviated or amputated.\\\\ 
Output only the class name and no additional text. Example: `Tablets'\\

\textcolor{gray}{\small (Llamma only)} \textbf{Product class from the list above is:}

\end{internallinenumbers}
\end{tcolorbox}
\vspace*{-2mm}
\caption{This prompt displays the template for LLM classification. Lines 09-10 are used solely for Few-shot prompting. Lines 13-14 are added only in the Combined-Reason attack scenario, while Line 18 is added for the Llamma-2 model, as we observed that it requires further prompt engineering to model the task as a completion prompt for outputting a product class.}
\label{prompt_classification}
\end{figure*}

\end{document}